\pdfoutput=1

\documentclass[11pt]{article}

\usepackage[final]{coling}

\usepackage{times}
\usepackage{latexsym}

\usepackage[T1]{fontenc}

\usepackage[utf8]{inputenc}

\usepackage{microtype}

\usepackage{inconsolata}

\usepackage{graphicx}

\usepackage{multicol}
\usepackage{booktabs}
\usepackage{makecell}
\usepackage{amsmath, amssymb}
\usepackage{multirow}
\usepackage{mathtools}
\usepackage{xcolor}
\usepackage{enumitem}
\usepackage{float}
\usepackage{graphicx}
\usepackage{upgreek}
\usepackage{seqsplit}
\usepackage{color,soul}
\usepackage{arydshln}
\usepackage{placeins}
\usepackage{pifont}
\usepackage{amssymb}
\usepackage{bbm}

\title{Embedding-Informed Adaptive Retrieval-Augmented Generation of \\ Large Language Models} 

\author{
 \textbf{Chengkai Huang\textsuperscript{1}},
 \textbf{Yu Xia\textsuperscript{2}},
 \textbf{Rui Wang\textsuperscript{3}},
 \textbf{Kaige Xie\textsuperscript{4}},
 \textbf{Tong Yu\textsuperscript{5}},
 \textbf{Julian McAuley\textsuperscript{2}},
 \textbf{Lina Yao\textsuperscript{1,6}}
\\
\\
 \textsuperscript{1}University of New South Wales,
 \textsuperscript{2}University of California San Diego,
 \textsuperscript{3}Duke University,
\\
 \textsuperscript{4}Georgia Institute of Technology,
 \textsuperscript{5}Adobe Research,
 \textsuperscript{6}CSIRO's Data61
\\
 \small{\textbf{Correspondence:} \href{chengkai.huang1@unsw.edu.au}{chengkai.huang1@unsw.edu.au}}
}

\begin{document}
\maketitle
\begin{abstract}

Retrieval-augmented large language models (LLMs) have been remarkably competent in various NLP tasks.
However, it was observed by previous works that retrieval is not always helpful, especially when the LLM is already knowledgable on the query to answer.
Motivated by this,
Adaptive Retrieval-Augmented Generation (ARAG) studies retrieving only when the knowledge asked by the query is absent in the LLM. 
Previous works of ARAG either require accessing the pre-training corpus or prompting with additional model inferences.
Aiming to avoid such drawbacks, we propose to determine whether the model is knowledgeable on a query via inspecting the (contextualized) pre-trained token embeddings of LLMs. 
We hypothesize that such embeddings capture rich information on the model's intrinsic knowledge base, which enables an efficient way of judging the necessity to retrieve from an external corpus.
Extensive experiments demonstrate our ARAG approach's superior performance across various benchmarks.
\end{abstract}


\section{Introduction}

Retrieval-augmented LLMs have exceled in various NLP tasks \cite{Li2022Survey, Yasunaga2023multimodal, LinTMT022}. 
However, it was observed \cite{MallenAZDKH23} that the retrieved knowledge might not necessarily improve the quality of a generated response, especially when the model is already knowledgeable on the input query from pre-training.
Moreover, the retrieval process can incur additional computational costs and latency, \emph{e.g.}, by significantly increasing the context length.

To solve this, Adaptive Retrieval-Augmented Generation (ARAG) dynamically determines whether the LLM has already acquired the knowledge to answer the query during pre-training, then only to retrieve from external corpus when the knowledge required is absent. The pilot work of ARAG  \cite{MallenAZDKH23} only retrieves if an entity from the query has low frequency in the pre-trained corpus. The LLMs are deemed not knowledgeable on the query if the extracted entities have low frequencies. An obvious drawback of such a heuristic approach is that it only works on \emph{entity-centric} question answering,  \emph{e.g.}, \textit{"What is the capital of Louisiana?"}, where the question is mostly about an entity (a person, country, etc.) that can be identified by existing entity extraction tools like Spacy \cite{spacy2}.
Additionally, the pre-trained corpus could be proprietary and not readily accessible to compute the frequency.
More recent approaches of ARAG, \emph{e.g.} \cite{RetrievalQA}, prompt the LLMs for a retrieval decision with extra LLM calls, assuming that the LLMs is aware of its knowledge boundary. This assumption is challenged by studies on LLM factuality, {\emph i.e.,} LLMs are overly confident in their ability to answer a question  \cite{ren2023investigating}.


In this work, we propose to decide whether to retrieve by analyzing the LLM pre-trained parameters, which does not require accessing the pre-trained data or extra LLM calls. Specifically, we predict whether the external knowledge from retrieval could help, by inspecting the pre-trained embeddings of tokens from the input query. We hypothesize that the pre-trained token embeddings
capture rich information on the model’s intrinsic knowledge base. This stems from previous analysis  \cite{cai2020isotropy}, claiming that tokens of different frequencies during pre-training tend to exhibit different layout in the embeddings space.  
In Section \ref{sec:embd}, we also show that the pre-trained embeddings are discriminative by the frequency information, consistent with  \cite{cai2020isotropy}.
Considering that such frequencies are indicative of whether the LLM is knowledgeable  \cite{MallenAZDKH23}, the token embeddings should contain sufficient information to judge the necessity to augment with external knowledge via retrieval. 
We term our approach \textbf{EI-ARAG} (Embedding-Informed ARAG).

\begin{figure*}[!t]
    \centering
        \includegraphics[width=0.3\linewidth]{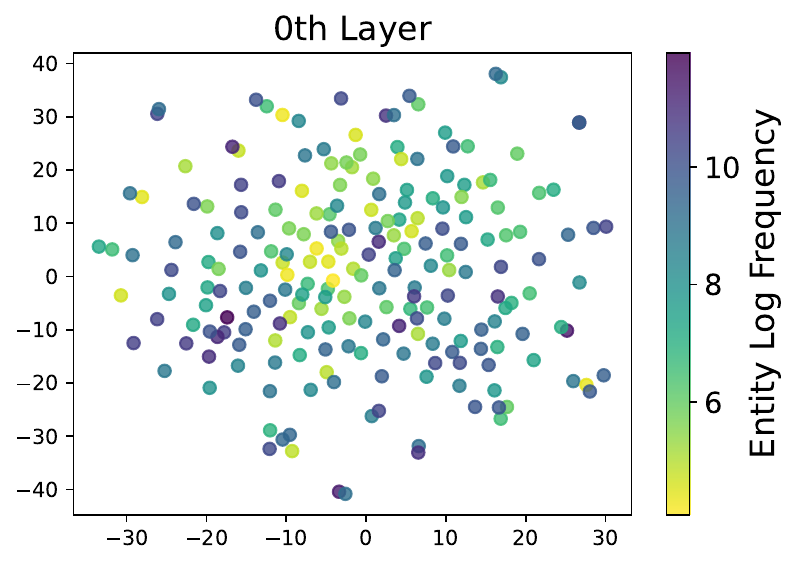}
        \includegraphics[width=0.3\linewidth]{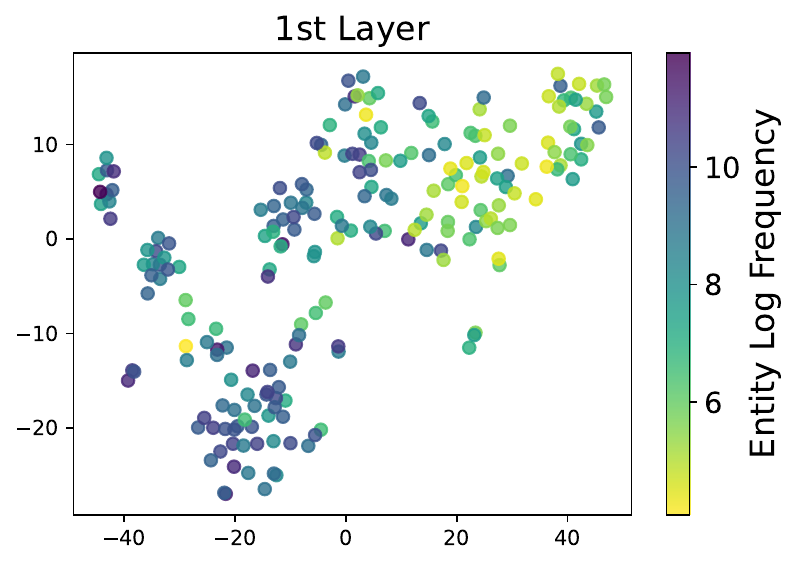}
        \includegraphics[width=0.3\linewidth]{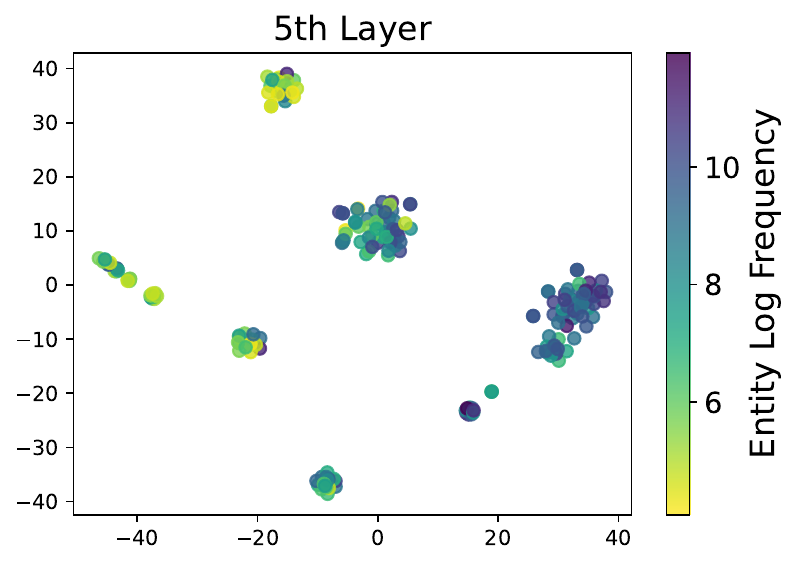}
        \includegraphics[width=0.3\linewidth]{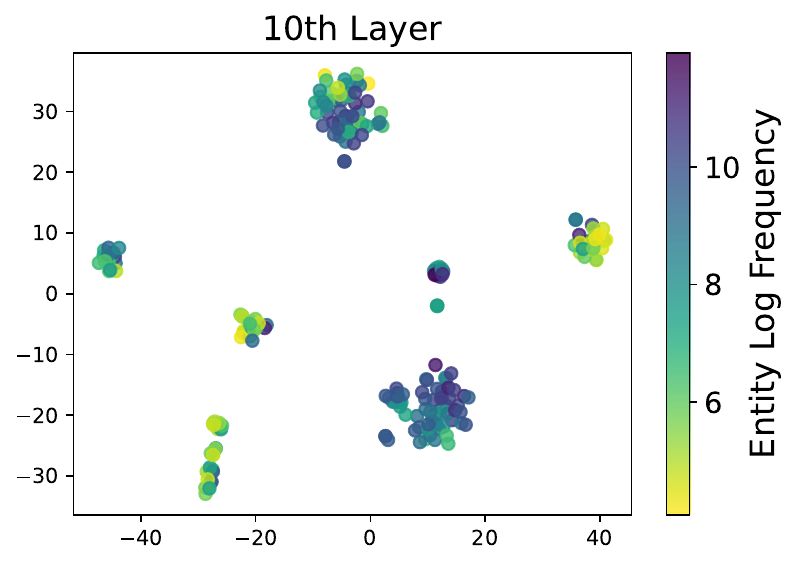}
        \includegraphics[width=0.3\linewidth]{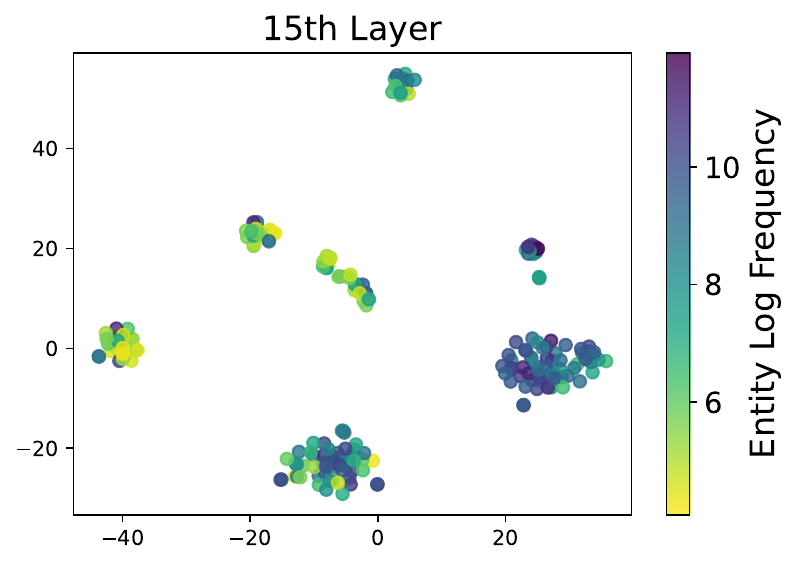}
        \includegraphics[width=0.3\linewidth]{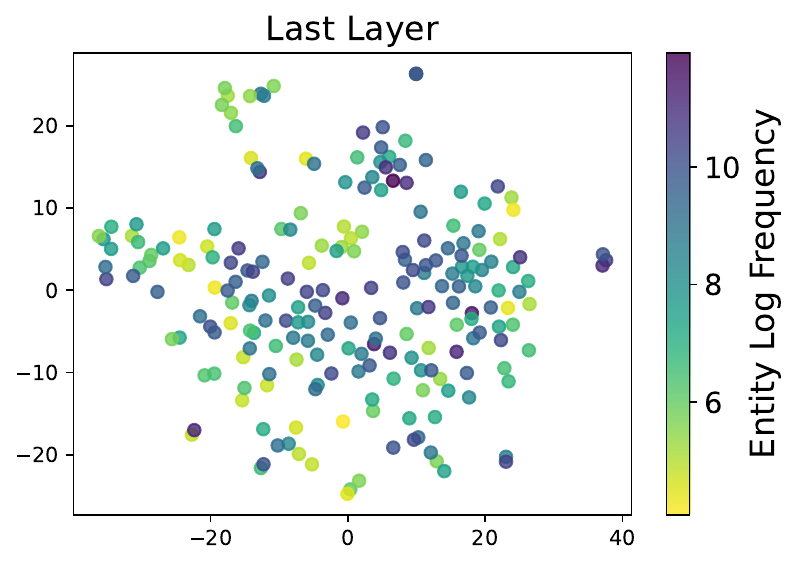}
    \caption{Visualization of embeddings at different layers of LLaMA 2 7B for director-related questions in PopQA. Darker color indicates the question contains entities with higher frequency in pre-training data.}
    \label{fig:embedding}
\end{figure*}

The main contributions of this work are as follows: 
\begin{itemize}

\item We hypothesis that the pre-trained LLMs token embeddings  capture rich information on the model's intrinsic knowledge base, and propose EI-ARAG to predict the necessity to retrieve based on these embeddings.

\item Instead of requiring extra LLM calls to judge whether the LLM is knowledgeable on the input, our proposed EI-ARAG shows that we can do it efficiently with the pre-trained token embedding. 

\item Extensive experiments and in-depth analyses demonstrate the superiority of our approach on various benchmarks.
\end{itemize}

To determine when to retrieve, existing methods either need to access pre-training data to calculate entity frequency (e.g., how many times `Louisiana' appears in pre-training corpus
\cite{MallenAZDKH23}), or require extra inference prompting LLMs to make decision themselves (e.g., ``Do you need to retrive external knowledge to answer the question correctly?'' \cite{RetrievalQA}). 
However, (i) relying on the entity frequency restricts the approach, making it suitable only for entity-centric question answering, (ii) the access to the pre-training data is not often available and (iii) the extra prompting will double the computational cost due to and additional LLM call.
To circumvent the above drawbacks, we hypothesis that the pre-trained token embeddings capture rich information on the model’s intrinsic knowledge (Section \ref{sec:embd}), proposing to utilize such embeddings to efficiently judge the necessity to retrieve from external knowledge base.




\section{Embedding-Informed Retrieval}
\label{sec:method_token}
While the input embedding layer (i.e.1st layer) contains rich semantic information of each token, later layers of LLMs provide embeddings that are more contextualized on current question. 
To capture more precisely LLMs' intrinsic knowledge given a specific question, we extract embeddings for each token from the contextualized layers.
We then obtain a sentence embedding for the given question via average pooling following \cite{Bohan2020}. 


Formally, given a question $q\sim Q$, we tokenize it with tokenizer $T$ and subsequently convert into a token embedding sequence, $\text{\textsf{embed}}_{\text{1st}}(\text{\textsf{T}}(q))$. 
We hypothesize that ${\textsf{embed}}_{\text{1st}}(\text{\textsf{T}}(q))$ encapsulates intrinsic knowledge of LLMs on the question $q$. To predict whether the LLM is already knowledge on $q$, we train a classifier $C: Q\rightarrow Y\in\{0,1\}$ whose prediction is $y\in Y$, where $y=1$ indicates the need for retrieval augmentation, and $y_i'=0$ indicates otherwise. Specifically, we have,
$$y = \text{\textsf{C}}(\text{\textsf{embed}}_{\text{1st}}(\text{\textsf{T}}(q))).$$
where $C$ is a neural network. We denoted its training set as $D=\{q_i, y_i\}_{i=1}^N$. The question $q_i$ is randomly sampled from the original training split of datasets. The label $y_i$ is determined by inferencing twice with the LLM on $q_i$, one with retrieval and one without retrieval. We determine $y_i=1$ if the result with retrieval is better than not. During the test, we only retrieve when $C$ predicts $1$.

To validate our hypothesis that the pre-trained embeddings contains rich information on the intrinsic knowledge, we visualize in Figure \ref{fig:embedding} the embeddings of entities in different layers of LLaMA 2 7B and color them based on the logarithmic entity frequency associated in each question following \cite{Xingyu2021}. 
We can observe that at the 0th layer, i.e., the input embedding layer, the embeddings is less discriminative in terms of the frequency. While on higher layers, we observe clear patterns for questions with entities of different frequency, which shows that the (contextualized) pre-trained embeddings captures the entity frequency information during pre-training, thus show be indicative on whether the LLM is knowledge about the given question \cite{MallenAZDKH23}.
Such embeddings-frequency correlation are also observed in prior analysis \cite{cai2020isotropy}.
While the embedding patterns appear from the first layer, embeddings from later layers do not bring obvious improvement to our classifier as shown in Table \ref{tab:layer_exp}. Thus, we simply extract sentence embeddings from the first contextualized layer to ensure computational efficiency compared to prompt-based methods that require full inference. More details are provided in Appendix \ref{sec:embd}.
Utilizing only the first contextualized layer, our embedding-informed method offers an efficient way to determine the need for retrieval augmentation compared to methods that need full inference \cite{RetrievalQA}. 
In contrast to the method requiring the access of the pre-training data \cite{MallenAZDKH23}, our method focuses on the analysis of embeddings, enhancing the applicability and scalability in real-world systems.

\begin{table*}[ht]
\tiny
\caption{Experiment results on the entity-centric PopQA dataset queries with different LLMs. We emphasize our results in bold for easy comparisons. \textit{ACC(\%)} denotes retrieval accuracy and \textit{POR(\%)} denotes the percentage of retrieval. Please refer to Section~\ref{sec:evame}.}
\label{tab:main}
\centering
\resizebox{\textwidth}{!}{
\renewcommand{\arraystretch}{1.1}
\begin{tabular}{llccccccccccccccc}
\toprule

& & \multicolumn{2}{c}{\bf GPT-Neo (1.3B)} & \multicolumn{2}{c}{\bf  GPT-Neo (2.7B)} & \multicolumn{2}{c}{\bf LLaMA 2 (7B)}  \\

\cmidrule(l{2pt}r{2pt}){3-6} \cmidrule(l{2pt}r{2pt}){7-12} \cmidrule(l{2pt}r{2pt}){13-17}

\textbf{Types} & \textbf{Methods} & \textbf{ACC(\%)}  & \textbf{POR(\%)}  & \textbf{ACC(\%)}  & \textbf{POR(\%)}   & \textbf{ACC(\%)}  & \textbf{POR(\%)}   \\
\midrule

\multirowcell{2}[-0.0ex][l]{Simple}

& No Retrieval & 11.19 & 0 & 12.56 & 0 & 24.64 & 0  \\

& Full Retrieval & 20.44 & 100 & 22.99 & 100 & 29.55 & 100 \\

\noalign{\vskip 0.25ex}\cdashline{1-17}\noalign{\vskip 0.75ex}

\multirowcell{4}[-0.0ex][l]{Adaptive} 

& DARAG~\cite{MallenAZDKH23} & 20.32 & 95.93 & 22.76 & 94.89 & 31.99 & 69.80  \\

& PARAG-Vanilla~\cite{RetrievalQA} & 11.85 & 0 & 12.56 & 0 & 27.78 & 88.98  \\

& PARAG-TAARE~\cite{RetrievalQA} & 11.85 & 0 & 12.56 & 0 & 29.21 & 95.15  \\





& \textbf{EI-ARAG (Ours)} & \textbf{21.47} & \textbf{84.33} & \textbf{24.00} & \textbf{94.37} & \textbf{33.08} & \textbf{57.89}  \\

\noalign{\vskip 0.25ex}\cdashline{1-17}\noalign{\vskip 0.75ex}

Oracle
& Adaptive-RAG w/ Oracle & 22.34 & 88.81 & 24.87 & 87.44 & 37.62 & 75.36 \\

\bottomrule
\end{tabular}
}
\vspace{-0.1in}
\end{table*}

\section{Experimental Setup}
In this section, we describe the datasets, baselines, metrics, and implementation details of our experiments. 
The experiments are designed to answer the following research questions:

\textbf{RQ1:} Can our method effectively decide when to retrieve for ARAG? How does it perform compared to state-of-the-art baselines? 

\textbf{RQ2:} Is using embeddings more computationally efficient than prompting-based methods?

\textbf{RQ3:} What kinds of knowledge are captured from the embeddings to  decide when to retrieve? The extensive discussion in Section \ref{sec:embd} has already answered RQ3 where we find contextualized embeddings to reflect LLM instrinsic knowledge similarly as entity frequency.

\textbf{Datasets.} Following the evaluation in \cite{MallenAZDKH23,RetrievalQA}, we verify the effectiveness of our proposed framework on two open-domain QA datasets (Entity QA dataset and Non-entity QA dataset) to answer our research questions. Please refer to Appendix \ref{sec:datasets} for details.
\textbf{Baselines.} We compare our EI-ARAG against relevant baselines, 
including: 
(1) \textbf{Simple Full Retrieval} \cite{MallenAZDKH23};(2) \textbf{DARAG} \cite{MallenAZDKH23}; (3) \textbf{PARAG-Vanilla} \cite{RetrievalQA}; (4)\textbf{PARAG-TAARE} \cite{RetrievalQA}; (5) \textbf{Adaptive-RAG w/ Oracle}: using the Oracle classifier with our Adaptive-RAG, which includes the best performance (i.e., the upper-bound performance) in an ideal scenario. See Appendix \ref{sec:baselines} for more details.

\textbf{Evaluation Metrics.}\label{sec:evame}
Following similar settings as \cite{MallenAZDKH23,RetrievalQA}, we report the results with (i) retrieval \textit{Accuracy (ACC)}, which evaluates how well LLMs can perform adaptive retrieval and (ii) \textit{Percentage of Retrieval (POR)} which assesses the efficiency of implementing adaptive retrieval.
\textit{Accuracy} measures the proportion of the predicted answer containing the ground-truth answer.  \textit{Percentage of Retrieval} quantifies the proportion of instances in which ARAG methods opt to activate retrieval across all test samples.

\section{Experimental Results}

\textbf{Entity-centric QA (RQ1).}\label{sec5.1} As shown in Table~\ref{tab:main}, we can observe that: 
(i) The simple No Retrieval method yields the lowest accuracy, highlighting the importance of incorporating external knowledge for enhanced response quality. Conversely, Full Retrieval shows a significant improvement in accuracy, affirming the benefit of external data. Among adaptive retrieval methods, DARAG, which is a prompting-based method, shows varied performance. 
Unfortunately, since GPT-Neo is not a chat model, DARAG tends to default to full retrieval on this platform, not achieving the desired adaptive retrieval effect. This limitation is evident as it performs well on the LLaMA 2 model but fails on the GPT-Neo models.
(ii) Our proposed EI-ARAG approach stands out by not only achieving high accuracy but also maintaining lower retrieval rates. This indicates efficient use of external information, retrieving only when the LLM lacks the necessary knowledge for a query. To help answer RQ3, in Table \ref{tab:layer_exp} and Figure \ref{fig:main_results}, we also run additional experiments and report the results on  PopQA, with varying layers when extracting the latent embeddings from LLaMA 2 7B.

\begin{table}[t!]
\small
\caption{Experiment results on the non-entity centric TriviaQA dataset queries with LLaMA 2 (7B). We emphasize our results in bold for easy comparisons. \textit{ACC(\%)} denotes retrieval accuracy and \textit{POR(\%)} denotes the percentage of retrieval. Please refer to Section~\ref{sec:evame}.}
\label{tab:sentence_exp}
\centering
\resizebox{0.475\textwidth}{!}{
\renewcommand{\arraystretch}{1.2}
\renewcommand{\tabcolsep}{2.5mm}
\begin{tabular}{lcc}
\toprule

\textbf{Methods} & \textbf{ACC (\%)} & \textbf{POR(\%)} \\
\midrule

\multirowcell{1}[-0.0ex][l]{No Retrieval} 
& 47.33 & 0 \\

\multirowcell{1}[-0.0ex][l]{Full Retrieval} 
& 62.33 & 100 \\

\noalign{\vskip 0.25ex}\cdashline{1-3}\noalign{\vskip 0.75ex}

\multirowcell{1}[-0.0ex][l]{DARAG~\cite{MallenAZDKH23}} 
& N/A & N/A \\

\multirowcell{1}[-0.0ex][l]{PARAG-Vanilla~\cite{RetrievalQA}} 
& 61.78 &  97.67 \\

\multirowcell{1}[-0.0ex][l]{PARAG-TAARE~\cite{RetrievalQA}} 
& 62.33 & 98.56 \\


\multirowcell{1}[-0.0ex][l]{\textbf{EI-ARAG (Ours)}}
& \textbf{62.67} & \textbf{92.11} \\
\noalign{\vskip 0.25ex}\cdashline{1-3}\noalign{\vskip 0.75ex}

\multirowcell{1}[-0.0ex][l]{Adaptive-RAG w/ Oracle}
& 68.56 & 52.67 \\

\bottomrule

\end{tabular}
}
\vspace{-0.125in}
\end{table}

\textbf{Non Entity-centric QA (RQ1).} \label{sec5.2}
In this subsection, we show the overall experimental results and offer in-depth analyses of our method in the non-entity-centric dataset TriviaQA. 

\begin{table}[ht]
\fontsize{6.5pt}{7pt}\selectfont
\centering
\setlength{\tabcolsep}{2pt} 
\renewcommand{\arraystretch}{1.1} 
\caption{Comparison of ARAG methods across different dimensions in LLaMA2 7B model (detailed in Section~\ref{sec5.2} and~\ref{sec5.3}). 
}
\begin{tabular}{>{\centering\arraybackslash}p{2cm} >{\centering\arraybackslash}p{2cm} >{\centering\arraybackslash}p{2cm} >{\centering\arraybackslash}p{1cm}}
\toprule
& \makecell{DARAG\\\cite{MallenAZDKH23}} & \makecell{PARAG\\\cite{RetrievalQA}} & \makecell{EI-ARAG\\(Ours)} \\
\midrule
\textbf{Entity QA} & \textcolor[HTML]{32CD32}{\ding{52}} & \textcolor[HTML]{32CD32}{\ding{52}} & \textcolor[HTML]{32CD32}{\ding{52}} \\
\textbf{Non-entity QA} & \textcolor{red}{\ding{55}} & \textcolor[HTML]{32CD32}{\ding{52}} & \textcolor[HTML]{32CD32}{\ding{52}} \\
\textbf{Latency (ms)} & 1740  & 2328 & 1744 \\
\textbf{ACC(\%)} & 31.99  & 29.21 & 33.08 \\
\textbf{POR(\%)} & 69.80 & 95.15 & 57.89 \\
\bottomrule
\end{tabular}
\label{tab:intro}
\end{table}

The main results as shown in Table~\ref{tab:sentence_exp} and the error analysis of our EI-ARAG are shown in Figure~\ref{fig:sankey}.
It is worth noting that DARAG is not suitable for sentence-level experiments because it is an entity-centric method, and the TriviaQA dataset contains questions involving multi-hop entities, resulting in an N/A outcome for DARAG. 
We observe that EI-ARAG is better than all baselines both for ACC and POR. Our approach reveals higher confidence in the LLM’s intrinsic knowledge than PARAG, resulting in fewer retrievals and more precise answers, demonstrating our method's superior decision-making on when to engage retrieval compared to the baseline model.





\textbf{Comparison of Computational Cost to Determine When to Retrieve (RQ2).}\label{sec5.3}
To better demonstrate the efficiency of our methods, we record the inference time needed by our methods to extract the first layer embedding of LLaMA 2 (7B).
In comparison, we also record the time needed for TAARE \cite{RetrievalQA} to prompt LLaMA 2 to generate ``Yes'' or ``No'' for retrieval.
We use the same instruction prompt as \citet{RetrievalQA} and limit the maximum generated new tokens to 5 to avoid extra inference cost.
On the test sets of PopQA and TrivialQA, our method extracts the first layer embeddings with only 0.0443 seconds on average for each question.
For the prompting-based method, it takes an average of 0.3885 seconds per question for the LLM to answer whether it needs retrieval.
The results further demonstrate the advantage of our methods for more efficient adaptive RAG.
To summarize, our method achieves the highest accuracy and the lowest proposition retrieval, while maintaining a relatively shorter average inference time. 

\section{Conclusion}
In this work, we introduce a novel approach for Adaptive Retrieval-Augmented Generation (ARAG), which strategically employs retrieval only when the LLM lacks the necessary knowledge for a query. Unlike previous methods that require access to the pre-training corpus or additional model inferences, our method leverages the rich information captured by the pre-trained token embeddings of LLMs. This approach allows us to efficiently determine the necessity of external retrieval. Extensive experiments demonstrate that our proposed ARAG method achieves superior performance across various benchmarks.

\section{Limitations}
The effectiveness of retrieval-augmented strategies depends on the quality and relevance of the external data sources. If the retrieved content is of low quality or irrelevant, it not only fails to aid in answering queries but might also introduce noise that can mislead the model and impair its decision-making process. Ensuring that retrieval mechanisms access high-quality and pertinent information is crucial for enhancing model performance and reliability. While this is out of the scope of this paper, future works are required to make RAG systems more robust and effective.

\section{Ethics Statement}
It is important to note that LLMs can still generate incorrect (hallucination) or biased outputs, even when they are retrieval-augmented. Therefore, it is always important to verify the outputs of language models with other sources of information.

\bibliography{custom}

\appendix
\newpage

\section{Related Work}
\paragraph{Retrieval-Augmented Generation.}
Retrieval-augmented Large Language Models (RALM) have been remarkably competent in various NLP tasks \cite{Li2022Survey}, which augments the input space of LMs with retrieved text passages \cite{GuuLTPC20}, leading to large improvements in knowledge-intensive tasks \cite{IzacardCHRBJG22}
However, standard RAG methods indiscriminately retrieve information regardless of the query, potentially degrading performance and increasing costs, as LLMs can often handle straightforward queries on their own, but noisy retrieved context might impede their performance.
To alleviate the limitations of RAG mentioned above, recent studies advocate for Adaptive RAG (ARAG), which dynamically determines retrieval necessity and relies only on LLMs’ parametric knowledge when considered unnecessary \cite{MallenAZDKH23}. Mallen et al. \cite{MallenAZDKH23} propose to heuristically retrieve when the popularity of an entity on Wikipedia is below a certain threshold. Jiang et al. \cite{JiangXGSLDYCN23} trigger retrieval if any token in the temporarily generated sentence has low confidence. \cite{Zhangyin,self-rag,Ruiyang} directly prompt LLMs for retrieval decisions, given the observation that LLMs can acknowledge their knowledge boundaries to some extent \cite{Zhangyue,Saurav}. Recently, Zhang et al. \cite{RetrievalQA} propose the TA-ARE, a simple yet effective method to help LLMs assess the necessity of retrieval, obviating the need for calibration or additional training.

\paragraph{Parametric and Non-parametric Knowledge.} 
Previous research demonstrates that large pre-trained language models (LMs) such as BERT, and GPT \cite{Jun2019, Bohan2020, Xingyu2021} encapsulate a significant amount of world knowledge within their parameters. However, depending exclusively on their parameters to store extensive world knowledge necessitates an impractically large number of parameters, and the knowledge can quickly become outdated. Recent studies \cite{MallenAZDKH23} show that enhancing LMs with non-parametric memories (i.e., retrieved text chunks) allows for smaller models to achieve performance levels comparable to those of larger models.

\begin{figure*}[ht]
    \centering
    \small
    \resizebox{0.75\linewidth}{!}{ 
        \includegraphics{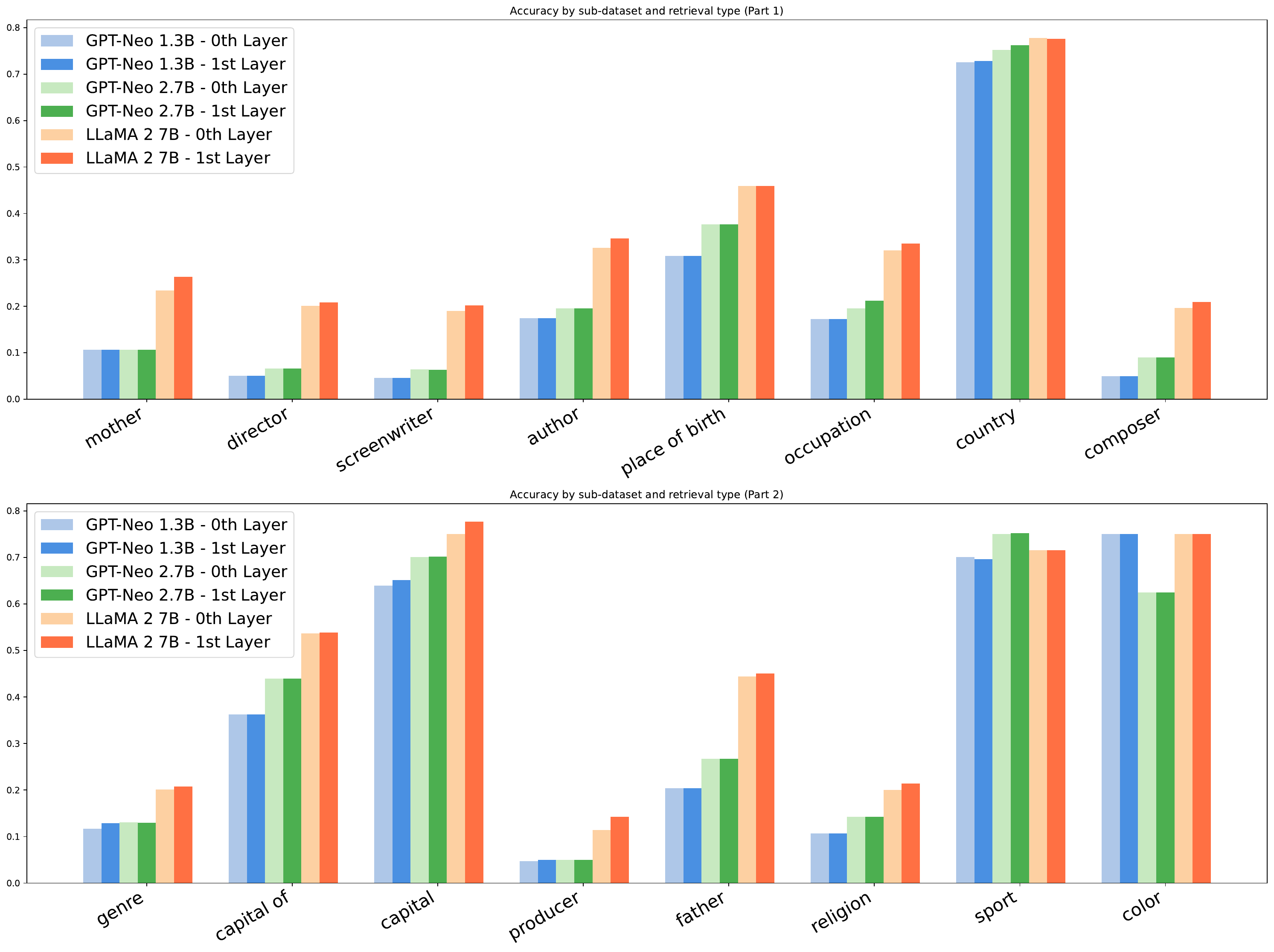}
    }
    \caption{Per-relationship type results on PopQA by different models, showing overall accuracy of EI-ARAG using 0th and 1st layer embeddings based on BM25 RALM.}
    \label{fig:main_results}
\end{figure*}

\section{Embedding-Informed ARAG}\label{sec:method}



In the context of open-domain QA, the primary objective of the ARAG method is to ascertain whether a given question (e.g., ``\textit{What is the capital of Louisiana?}'') requires retrieval augmentation.
The core of this task is to determine whether language models already possess knowledge related to the question, thereby deciding if there is a need to retrieve external knowledge bases to enhance the model prediction.  
This adaptive retrieval approach can effectively save context length and thus reduce latency during LLM inference. 
Besides, it can also mitigate performance degradation caused by redundant retrievals in LLMs \cite{MallenAZDKH23}.

\section{Why the Pre-Trained Embeddings Help?}\label{sec:embd}
To validate our hypothesis that the pre-trained embeddings contain rich information on the intrinsic knowledge and are sufficient to help judge the necessity to retrieve, 
we visualize in Figure \ref{fig:embedding} the embeddings of entities in  different layers of LLaMA 2 7B 
and color them based on the logarithmic entity frequency associated with each question following \cite{Xingyu2021}. 
Questions are the entire director-related subset of PopQA dataset \cite{MallenAZDKH23} selected as a representative challenging subset.

\begin{table}[t!]
\caption{EI-ARAG accuracy with different embedding layers of LLaMA 2 7B averaged on all subsets of PopQA.}
\small
\centering
\renewcommand{\arraystretch}{1}
\renewcommand{\tabcolsep}{2.5mm}
\begin{tabular}{lc}
\toprule

\textbf{Methods} & \textbf{ACC (\%)} \\
\midrule

\multirowcell{1}[-0.0ex][l]{0th Layer} 
& 38.54  \\

\multirowcell{1}[-0.0ex][l]{1st Layer} 
& 40.98 \\

\multirowcell{1}[-0.0ex][l]{5th Layer} 
& 40.97 \\

\multirowcell{1}[-0.0ex][l]{10th Layer} 
& 41.13 \\

\multirowcell{1}[-0.0ex][l]{15th Layer} 
& 41.09 \\

\multirowcell{1}[-0.0ex][l]{Last Layer} 
& 40.86 \\

\bottomrule

\end{tabular}
\label{tab:layer_exp}
\vspace{-0.2in}
\end{table}

From Figure \ref{fig:embedding}, we can observe that at the 0th layer, i.e., the input embedding layer, the embeddings is less discriminative in terms of the frequency.
While on higher layers, we observe clear patterns for questions with entities of different frequencies.
In Table \ref{tab:layer_exp}, we show the testing results of using the embeddings from different layers.
We find that the 0th layer is also producing detect results.
Therefore, we speculate that embeddings from the 0th layer should also be correlated with frequency, but require a more complex visualization method to exhibit in a 2D plot.
The results in Figure \ref{fig:embedding} show that the (contextualized) pre-trained embeddings capture the entity frequency information during pre-training, thus indicative of whether the LLM is knowledgeable about the given question \cite{MallenAZDKH23}.
Such embeddings-frequency correlations are also observed in prior analysis  \cite{cai2020isotropy}.

While the embedding patterns appear from the first layer, embeddings from later layers do not bring obvious improvement to our classifier as shown in Table \ref{tab:layer_exp}.
Thus, in our method, we simply extract sentence embeddings from the first contextualized layer to ensure computational efficiency compared to prompt-based methods that require full inference.
We also show in Figure \ref{fig:main_results} additional results on the full PopQA dataset comparing the performance of utilizing the 0th layer and 1st layer in Table \ref{tab:layer_exp}.

\section{Baselines}\label{sec:baselines}

We compare our EI-ARAG against relevant baselines, including two retrieval-augmented LLM strategies (No retrieval and full retrieval) and the adaptive retrieval approaches \cite{MallenAZDKH23,RetrievalQA}, which can be grouped into one of two categories: Simple and Adaptive. Specifically, baseline approaches include: 
\textbf{Simple No Retrieval}: Directly input the question prompt into LLMs to generate answers. 
\textbf{Simple Full Retrieval} \cite{MallenAZDKH23}: For all questions, we use the retriever to retrieve related external knowledge, which is then added to the prompts before inputting them into large language models to generate answers.
\textbf{DARAG} \cite{MallenAZDKH23}: Data-aware Adaptive RAG determines the complexity level questions based on the frequency of its entities. It suggests using the retrieval modules only when the entity frequency falls below a certain threshold. In the PopQA dataset, this threshold for each relation type is determined using brute force search.
\textbf{PARAG-Vanilla} \cite{RetrievalQA}: Prompting Adaptive RAG (Vanilla) uses vanilla prompt to ask LLMs whether retrieval is necessary to determine ``Yes'' or ``No''. If retrieval is needed, proceed with it; if not, do not perform retrieval. This adaptive retrieval method requires two complete inferences.
\textbf{PARAG-TAARE} \cite{RetrievalQA}: Prompting Adaptive RAG (Time-aware) incorporates time-related prompts to ask LLMs whether retrieval is needed to determine ``Yes'' or ``No''. If retrieval is required, it will be conducted; otherwise, it will not. This adaptive retrieval method also requires two complete inferences.
\textbf{Adaptive-RAG w/ Oracle}: using the Oracle classifier with our Adaptive-RAG, which includes the best performance (i.e., the upper-bound performance) in an ideal scenario. 
In the ideal setting, Adaptive approaches are expected to be more effective than those using Full Retrieval, while also being more efficient than the No Retrieval approach.

For the fair comparison, we evaluate our framework across models of varying capacities: GPT-Neo (1.3 billion), GPT-Neo (2.7 billion) \cite{GPT-Neo}, LLaMA 2 (7 billion) \cite{Llama2}. In this evaluation, we do not perform any fine-tuning on models. 

\section{Dataset Construction}\label{sec:datasets}

\textbf{PopQA} \cite{MallenAZDKH23}: An entity-centric open-domain QA dataset about entities with a wide variety of popularity, consisting of 14k questions to cover factual information in the long tail that might have been missed in other popular QA datasets \cite{KwiatkowskiPRCP19}. Following the setting of \citet{MallenAZDKH23}, we choose 75\% of data for training, then evaluate our model and baselines on the remaining 25\% of PopQA.

\textbf{TriviaQA} \cite{JoshiCWZ17}: This Non-entity QA dataset is designed for reading comprehension tasks and consists of question-answer-evidence triples, where each question incorporates both entity and non-entity information, challenging models to navigate and integrate diverse types of data within a single query, thus providing a more complex test of the models' comprehension and reasoning abilities.
Since the TriviaQA-unfiltered (open) test set is not publicly available, we derived from a random sampling of 3,600 entries from the original TriviaQA-unfiltered dataset, with a training and testing split of 2,700 and 900, respectively. For the RAG task, we utilize the top-5 documents retrieved from Google Search API to gather relevant passages.

\section{Details of Data Constructions}
\label{sec:appendix}

In this paper, for entity-centric experiments, we adopt the 16 relationship types identified in the PopQA dataset as outlined in  \cite{MallenAZDKH23}. The authors of this reference manually annotated templates to transform knowledge triples into natural language questions. The complete list of templates utilized to generate EI-ARAG is presented in Table~\ref{tab:list_of_instructions_ours}.

\begin{table}[h!]
\footnotesize
\renewcommand{\arraystretch}{1.2}
\setlength{\tabcolsep}{2pt}
    \centering
    \begin{tabular}{lr}
\toprule
\textbf{Relationship } & \textbf{Template} \\\midrule
occupation  & What is \texttt{[subj]} 's occupation? \\
place of birth  & In what city was \texttt{[subj]} born? \\
genre  & What genre is \texttt{[subj]}?  \\
father  &  Who is the father of \texttt{[subj]} ?\\
country & In what country is \texttt{[subj]} ? \\
producer & Who was the producer of \texttt{[subj]} ? \\
director & Who was the director of \texttt{[subj]} ? \\
capital of & What is \texttt{[subj]}  the capital of? \\
screenwriter & Who was the screenwriter for \texttt{[subj]} ?\\
composer & Who was the composer of \texttt{[subj]} ? \\
color & What color is \texttt{[subj]} ? \\
religion & What is the religion of \texttt{[subj]} ? \\
sport & What sport does \texttt{[subj]}  play? \\
author & Who is the author of \texttt{[subj]} ? \\
mother & Who is the mother of \texttt{[subj]} ? \\
capital &What is the capital of \texttt{[subj]} ? \\
\bottomrule
 \end{tabular}
    \caption{Full list of the manually annotated templates used for our creations. \texttt{[subj]} denotes a placeholder for subject entities. }\label{tab:list_of_instructions_ours}
\end{table}

\begin{table*}[ht]
\caption{Case study with LLaMA 2 (7B). We represent the factual error in red and the accurate information in blue. 
}
\vspace{-0.05in}
\label{tab:case}
\small
\centering
\resizebox{\textwidth}{!}{
\renewcommand{\arraystretch}{1.3}
\begin{tabular}{cccc}
\toprule

\multicolumn{1}{p{.12\textwidth}}{\textbf{Dataset}} & \multicolumn{1}{p{.178\textwidth}}{\textbf{Question}} & \multicolumn{1}{p{.4\textwidth}}{\textbf{PARAG-TAARE \cite{RetrievalQA}}} & \multicolumn{1}{p{.4\textwidth}}{\textbf{EI-ARAG (Ours)}} \\

\midrule
\multicolumn{1}{p{.12\textwidth}}{\textbf{PopQA} \quad \; }&
\multicolumn{1}{p{.178\textwidth}}{Who is the mother of \textcolor{blue}{Melissa Benn?}} &
\multicolumn{1}{p{.4\textwidth}}{\textbf{Query Type: } (Non Retrieval)\newline \textbf{Generation: } \textcolor{red}{Hilary Mantel}.} &
\multicolumn{1}{p{.4\textwidth}}{\textbf{Query Type: } (Retrieval)\newline 
\textbf{Retrieved Passages: } 'Melissa Ann Benn (born 1957) is a British journalist and writer. She is the only daughter of Tony and \textcolor{blue}{Caroline Benn}.'
\newline
\textbf{Generation: } \textcolor{blue}{Caroline Benn}.} \\ \midrule

\multicolumn{1}{p{.12\textwidth}}{\textbf{PopQA} \quad \; }&
\multicolumn{1}{p{.178\textwidth}}{What is \textcolor{blue}{Peter W. Barca}'s occupation?} &
\multicolumn{1}{p{.4\textwidth}}{\textbf{Query Type: } (Retrieval)\newline
\textbf{Retrieved Passages: } 'Fever/Dream is a play by Sheila Callaghan which premiered in 2009 at Woolly Mammoth \textcolor{red}{Theatre Company} in Washington, DC. It is a reinvention of Pedro Calderón de la Barca's play Life is a Dream.'
\newline
\textbf{Generation: } \textcolor{red}{actor}.} &
\multicolumn{1}{p{.4\textwidth}}{\textbf{Query Type: } (Non Retrieval)\newline 
\textbf{Generation: } \textcolor{blue}{politician}.} \\ \midrule \midrule

\multicolumn{1}{p{.12\textwidth}}{\textbf{TriviaQA}}&
\multicolumn{1}{p{.178\textwidth}}{Against which South American country was Bolivia fighting during 'The War of the Pacific?} &
\multicolumn{1}{p{.4\textwidth}}{\textbf{Query Type: } (Retrieval)\newline \textbf{Retrieved Passages: }Bolivia then declared war on Chile and called upon Peru for help. Chile declared war on both \textcolor{red}{Peru} and Bolivia (April 5, 1879).\newline \textbf{Generation: } \textcolor{red}{Peru}.}
 &

\multicolumn{1}{p{.4\textwidth}}{\textbf{Query Type: } (Non Retrieval)\newline \textbf{Generation: }Bolivia fought against \textcolor{blue}{Chile} during the War of the Pacific.} \\

\midrule

\multicolumn{1}{p{.12\textwidth}}{\textbf{TriviaQA}}&
\multicolumn{1}{p{.178\textwidth}}{Who was the first man to orbit the Earth?} &
\multicolumn{1}{p{.4\textwidth}}{\textbf{Query Type: } (Retrieval)\newline \textbf{Retrieved Passages: } \textcolor{red}{John Glenn}, the First American to Orbit the Earth aboard ...\newline \textbf{Generation: } \textcolor{red}{John Glenn}.}
 &

\multicolumn{1}{p{.4\textwidth}}{\textbf{Query Type: } (Non Retrieval)\newline \textbf{Generation: } The first man to orbit the Earth was \textcolor{blue}{Yuri Gagarin}, a Soviet cosmonaut.} \\

\bottomrule

\end{tabular}
}
\vspace{-0.125in}
\end{table*}

\section{Implementation Details}

For conducting experiments, we use a single NVIDIA RTX A5000 GPU equipped with 24GB of GPU memory. Besides, we use int8bit \cite{GLM130B} quantization with LLaMA2 7 billion to make them fit our GPUs. Similar to  \citet{MallenAZDKH23}, in our experiments with GPT-Neo 1.3 and 2.7 billion parameters, we did not detect a notable performance decline when employing quantization.
Besides, for a fair comparison, by configuring the number of beams as 1 and setting sampling to False, we ensured that the LLMs consistently produced fixed inference results following \cite{MallenAZDKH23}. 

For the prompts design, we follow the previous work's experiment setting  \cite{MallenAZDKH23}. For the PopQA dataset, we sample few-shot examples stratified by relationship type to diversify the samples: for each of the 15 relationship types other than the one in the test question, we sample one random question-answer pair to include in the context (detailed relation type templates can be found in Appendix A of  \cite{MallenAZDKH23}).
Besides, since the quality of the retrieved documents is not the focus of this paper, we use the off-the-shelf BM25 retrieval model and author-provided top-5 documents \cite{MallenAZDKH23} extracted from Wikipedia where possible.
For the TriviaQA dataset, we adopt the standard template following as  \cite{RetrievalQA}.
Consistency across experiments is maintained by employing the same random seed as used in  \cite{MallenAZDKH23}, guaranteeing that all models are tested against a standardized set of exemplars. Consequently, the 15 exemplars provided to each model are uniformly generated using this identical seed. Besides, all questions are formatted using the prompt template `Q: <question> A:' to facilitate generative prediction. A 15-shot prompting method is adopted to maintain uniformity across both the GPT-Neo and LLaMA2 models.
In our EI-ARAG, for the sentence classifier, we use and train a three-layer MLP, by considering a good tradeoff between the accuracy and latency. Specifically, the classifier is trained using the epoch that shows the best performance until 50 training iterations from the validation set,with the learning rate of 1e-3 and the Adam \cite{KingmaB14} as an optimizer.

\section{Error Analysis on TriviaQA}
\begin{figure}[htbp]
    \centering
    \includegraphics[width=\linewidth]{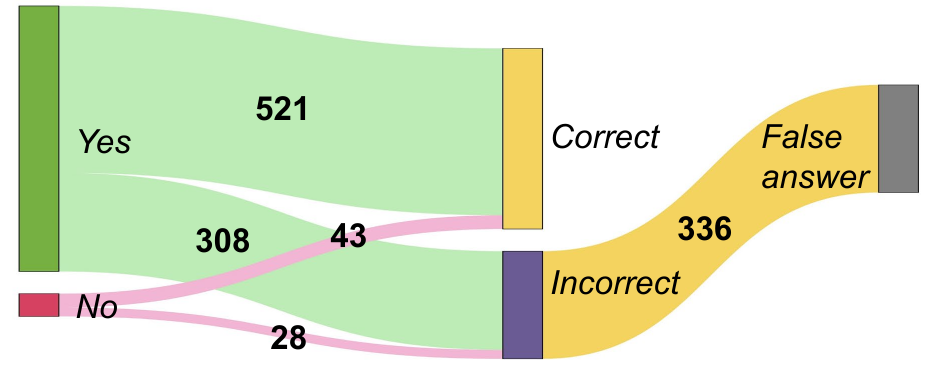}
    \caption{Sankey Diagram for our EI-ARAG method on the  TriviaQA dataset.}
    \label{fig:sankey}
\end{figure}

We plot the error analysis on LLaMA2 7B using 
our EI-ARAG in Figure~\ref{fig:sankey}, which illustrates the ARAG cases in TriviaQA. 
The green area indicates when ARAG models activate retrieval, whereas the red area shows when they assess no external information is needed. The yellow area is where retrieval may or may not occur but predictions are incorrect predictions. 

\section{Case Study}

We conduct a case study to qualitatively compare our Adaptive-RAG against previous Adaptive Retrieval \cite{RetrievalQA}. Table~\ref{tab:case} shows the 
classification
complexity and the query handling patterns for both entity-centric (PopQA) and sentence-level (TriviaQA) questions.

\paragraph{Example 1:} For this entity-centric question, TAARE Adaptive-RAG \cite{RetrievalQA} identifies that it is answerable by only using the LLM’s parametric knowledge about ‘Melissa Been' and generating the hallucination information.
In contrast, our model fetches additional documents, leading to producing correct responses about ‘Melissa Been'. 

\paragraph{Example 2:} In the sentence-level scenario involving historical conflict, PARAG-TAARE \cite{RetrievalQA} seeks out relevant information, including details like ‘Peru’, but incorrectly associates the event with 'Peru' due to an error in retrieving contextually relevant information. While our Adaptive Retrieval identifies this sentence knowledge have been stored in LLMs not need to request such information from external sources, resulting in accurate answers.

In the additional two examples in Table~\ref{tab:case} we have similar observations, which validate the advantages of our approach.

\end{document}